\documentclass[lettersize,journal]{IEEEtran}
\usepackage{amsmath,amsfonts}
\usepackage{algorithmic}
\usepackage{algorithm}
\usepackage{array}
\usepackage{textcomp}
\usepackage{stfloats}
\usepackage{url}
\usepackage{verbatim}
\usepackage{graphicx}
\usepackage{subfigure}
\usepackage{cite}
\usepackage{booktabs}

\begin{document}
	
	\title{Effective Multimodal Reinforcement Learning with Modality Alignment and Importance Enhancement}
	
	\author{Jinming Ma, Yingfeng Chen, Feng Wu,  Xianpeng Ji, and Yu Ding 
		\thanks{This work is supported in part by the Major Research Plan of the National Natural
			Science Foundation of China under Grant 92048301.}
		\thanks{Jinming Ma and Feng Wu are with School of Computer Science and Technology, University of Science and Technology of China, Hefei,
			China (e-mail: jinmingm@mail.ustc.edu.cn, wufeng02@ustc.edu.cn).}
		\thanks{Yingfeng Chen, Xianpeng Ji, and Yu Ding are with Netease Fuxi AI Lab, Hangzhou, China (e-mail: chenyingfeng1@corp.netease.com, jixianpeng@corp.netease.com, dingyu01@corp.netease.com)}
		\thanks{Feng Wu is the corresponding author.}
	}
	
	
	
	\maketitle
	
	\begin{abstract}
		Many real-world applications require an agent to make robust and deliberate decisions with multimodal information (e.g., robots with multi-sensory inputs). However, it is very challenging to train the agent via reinforcement learning (RL) due to the heterogeneity and dynamic importance of different modalities. Specifically, we observe that these issues make conventional RL methods difficult to learn a useful state representation in the end-to-end training with multimodal information. To address this, we propose a novel multimodal RL approach that can do multimodal alignment and importance enhancement according to their similarity and importance in terms of RL tasks respectively. By doing so, we are able to learn an effective state representation and consequentially improve the RL training process. We test our approach on several multimodal RL domains, showing that it outperforms state-of-the-art methods in terms of learning speed and policy quality.
	\end{abstract}
	
	
	\section{Introduction}
	
	\IEEEPARstart{D}{eep} reinforcement learning has made remarkable progress recently in many tasks, such as playing games \cite{mnih2015human} and controlling robots \cite{levine2016end,singh2019end}. Those works can train in the end-to-end fashion with raw sensory inputs, but to date mostly based on a single visual modality. Cognitive and psychology studies \cite{spelke1976infants} reveal humans are able to use multiple sources of information (e.g., vision, audio, and tactile) to build a better understanding of the physical world and make decisions. Inspired by this, multimodal machine learning (ML) aims to build models that can process information from multiple modalities in many applications \cite{ngiam2011multimodal,srivastava2012multimodal,hodosh2013framing,d2015review,kahou2016emonets,huang2020multi}. Generally, it is believed that multimodal information is crucial for agents to make robust and deliberate decisions. Multimodal RL becomes an active RL topic and many applications based on it \cite{liu2017learning,misra2017mapping,zhang2018multimodal,chaplot2019embodied,henkel2019score} have been successfully developed.
	
	Although multimodal information is indeed beneficial for an agent to make decisions, it also brings many challenges to RL algorithms. Firstly, the {\em heterogeneity} nature of multiple modalities makes it difficult to form a consistent representation in deep neural networks \cite{baltruvsaitis2018multimodal}. For example, to capture what a speaker said, the visual (i.e., articulation and muscle movements) and sound (i.e., amplitude and frequency) features must be always properly aligned. Modality alignment, one of the key challenges of general multimodal ML, may look straightforward for supervised learning with labeled data but is very challenging for RL due to indirect reward feedback \cite{de2018integrating}. When an agent is trained in an end-to-end manner, simply multimodal fusion actually slows down the RL training. We observed in our experiments that the slowdown is mainly due to unaligned multimodal inputs that cause issues for learning a state representation in RL.
	
	Secondly, modalities may play different {\em importance}, measured by information quantity, at some periods in dynamic environments. For example, self-driving cars often rely on different sensors in different weather conditions. Therefore, an agent should be able to dynamically bias towards more informative modalities and enhance the importance of themselves. We observed that RL without doing so will fail to learn an effective state representation and thereby good policy. To this end, previous work \cite{omidshafiei2017crossmodal} proposed to learn appropriate attention model to the required modality at the right moment to captures temporal crossmodal dependencies. Unfortunately, reasoning the importance of modalities and integrating them into state representation is a hard task, and mixing it with sequential decision-making in multimodal RL is even more challenging. Therefore, an attention model for this purpose may be very difficult to train in complex domains as shown later in our experiments.
	
	Against this background, we propose a novel multimodal RL with effective state representation learning, targeting at the {\em modal heterogeneity} and {\em dynamic importance} issues. Respectively, our approach consists of two main modules, i.e., {\em modality alignment} and {\em importance enhancement}. Specifically, in our modality alignment module, we jointly train the feature extractor of each modality to align with each other in the embedding space, based on some similarity measurement \cite{aytar2017see}. In our importance enhancement module, we compute the mean and variance of each modality given a sequence of data at training time and bias towards the modalities that deviate further from their means. This enhancement strategy is inspired by the fact that observations occurring with a lower probability\footnote{Note that observations occurring with high probability are usually clear enough for the agent to make correct decision and therefore enhancement is not necessary in this case.} are often more informative \cite{informationTheory}. By combining these two modules together, we are able to learn an effective state representation, which is the key to the performance of RL given multimodal information. In the experiments, we evaluate our approach on two benchmark domains for multimodal RL and a more realistic case study on self-driving car control with camera and lidar sensors. By comparing with several baselines and state-of-the-art methods, we confirm the effectiveness of our approach that can learn a policy more quickly with substantially higher quality. Moreover, we conducted empirical analysis on the modules to illustrate how the two main modules work and how the two key issues are addressed.
	
	\section{Related Work}
	
	\noindent In this section, we will review the related work from the following research threads: multimodal machine learning and multimodal reinforcement learning.
	
	\subsection{Multimodal Machine Learning}
	
	\noindent Multimodal machine learning aims to build models that can process and correlate information from multiple modalities to solve some multimodal applications. Early works in this field mainly include audio-visual speech recognition (AVSR) \cite{ngiam2011multimodal,  srivastava2012multimodal}, multimedia content indexing and retrieval \cite{atrey2010multimodal, vo2019composing}, multimedia description \cite{hodosh2013framing}, multimodal emotion recognition \cite{kahou2016emonets}, visual question-answering \cite{kim2016multimodal}, multimodal interaction with humans \cite{d2015review} and autonomous driving \cite{huang2020multi, xiao2019multimodal}. The research of multimodal in deep learning is quite popular and fruitful, which can offer many thoughts and methods for multimodal reinforcement learning. 
	
	However, applying these methods to reinforcement learning still has many aforementioned challenges, such as multimodal alignment, different importance of multiple modalities and heterogeneous modalities. 
	Specifically, RL is a sequential decision-making problem with the objective of maximizing accumulated rewards of multiple steps. A one-step decision in RL may have long-term some consequences, which are indirectly reflected by successive rewards when some key events happen (e.g., goal achieved). This is very different from machine learning where a direct correction can be made using the labels. 
	Note that multi-modal fusion in RL is much more challenging without direct feedback (i.e., whether it is a good fusion or not). We observed that the key to the success of multi-modal RL is to learn a good “state” representation. Here, the states have a specific meaning that must summarize all the necessary information to make a correct decision in MDP. In RL setting, the underlying state representation is hidden and must be learned using the gradient (in terms of rewards) from some typical RL algorithms (e.g., A2C). Canonical multi-modal ML approaches trained in this setting do not perform well, which will be demonstrated in experiment.
	
	\subsection{Multimodal Reinforcement Learning}
	
	\noindent Multimodal RL is an important area that is not well developed yet. It's rare but there is already some pioneering work try to combine multimodal learning and reinforcement learning, such as using multimodal reinforcement learning to solve end-to-end autonomous driving \cite{liu2017learning}, robot control \cite{chaplot2019embodied}, dialogue systems \cite{misra2017mapping, zhang2018multimodal}, and score following in an end-to-end fashion \cite{henkel2019score}. Other interesting work \cite{silva2019playing} explores the use of latent representations obtained from multiple input sensory, training policies over different subsets of input modalities. CASL~\cite{omidshafiei2017crossmodal} uses the Option-Critic framework and attentive mechanism to learn transferable skills across multiple sensory inputs. These methods have achieved certain success in specific scenarios. 
	
	However, most of these works have not comprehensively addressed the challenges mentioned in our work and the policy may be difficult to learn with multiple modalities. Specifically, the key challenge of RL with hidden state representation and indirect (usually delayed) feedback remains unsolved, which indeed makes multi-modal RL a unique problem. This motivates us to propose a new method that works better for multi-modal RL.
	
	\section{Background}
	
	\subsection{Formal Model}
	
	\noindent Here, we briefly introduce our settings of multimodal RL. Similar to standard RL, an agent interacts with the environment by taking actions under state and collecting rewards. The key difference is that the agent receives an observation consisting of multiple modalities rather than the true state (e.g., robots with multi-sensory inputs). Given this, the agent must be able to reason about the underlying state information from the multimodal observations in order to learn a proper policy. Therefore, we model our problem as a multimodal MDP.
	
	Specifically, the multimodal MDP can be formalized by the tuple $\left\langle S,A,T,R,\gamma  \right\rangle $, where: $S$ is the state space; $A$ is the action space; $T$ is the transition function; $R$ is the reward function; $\gamma$ is the discount factor.
	At each time step, we assume the agent receives an observation $ \mathbf{O} = (o^1, \cdots, o^k)$ of $k$ modalities for the underlying state $s$.
	Hence, the objective of multimodal RL is to learn an optimal policy $\pi^*$, mapping from a sequence of multimodal observations to an action.
	
	As detailed later, we build our approach upon the advantage actor-critic (A2C) method \cite{schulman2015high} for its simplicity and popularity, though generally any RL algorithms can be integrated with ours.
	Specifically, it learns policy by minimizing the actor and critic losses respectively:
	
	\begin{equation}
		\label{Loss-critic}
		\mathcal{L}(w) = \frac{1}{2|\mathcal{B}|} \sum_{t=1}^{|\mathcal{B}|} \left( \hat{R} - V_w(s_t) \right)^2
	\end{equation}
	\begin{equation}
		\label{Loss-actor}
		\mathcal{L}(\theta) = - \frac{1}{|\mathcal{B}|} \sum_{t=1}^{|\mathcal{B}|} \log \pi_{\theta}(a_t|s_t) A_t
	\end{equation}
	where $\mathcal{B}$ is the buffer, $A_t = \hat{R} - V_{w}(s_t)$ is the advantage and $\hat{R} = \sum_{t=1}^{T-1} \gamma^t r_t +\gamma^{T-1} V_{w}(s_T)$.
	
	Previous studies on multimodal RL are mostly on its applications in an {\em end-to-end} fashion, such as autonomous driving \cite{liu2017learning}, robot control \cite{chaplot2019embodied}, dialogue systems \cite{misra2017mapping,zhang2018multimodal}, and score following \cite{henkel2019score}. 
	Besides, \cite{silva2019playing} explores the use of latent representations obtained from multiple input sensory, training policies over different subsets of input modalities. \cite{omidshafiei2017crossmodal} uses the option-critic and attentive mechanism to learn transferable skills across multiple sensory inputs.
	
	Although multimodal RL is a natural choice for many real-world applications as aforementioned, little research has been done about the major impact of multimodal information on the training of RL agent, which motivates our work.
	
	\subsection{Motivation and Challenges}
	
	\noindent It is worth noting that multimodal RL shares several challenges with general multimodal machine learning \cite{baltruvsaitis2018multimodal}, as both of them take multimodal information as the inputs and try to learn a representation. Nevertheless, this task is especially challenging for multimodal RL because the states (i.e. representation goals) are usually hidden and must be learned implicitly from reward signals instead of directly from supervised labels. In this paper, we focus on the following two challenges that are observed to have major impact on the RL training process.
	
	
	\begin{figure}[t]
		\centering
		\subfigure[Gridworld domain.]
		{\includegraphics[width=0.38\hsize]{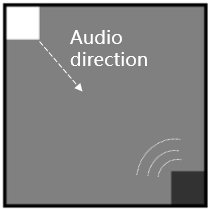}} 
		\subfigure[Traning curves.]
		{\includegraphics[width=0.6\hsize]{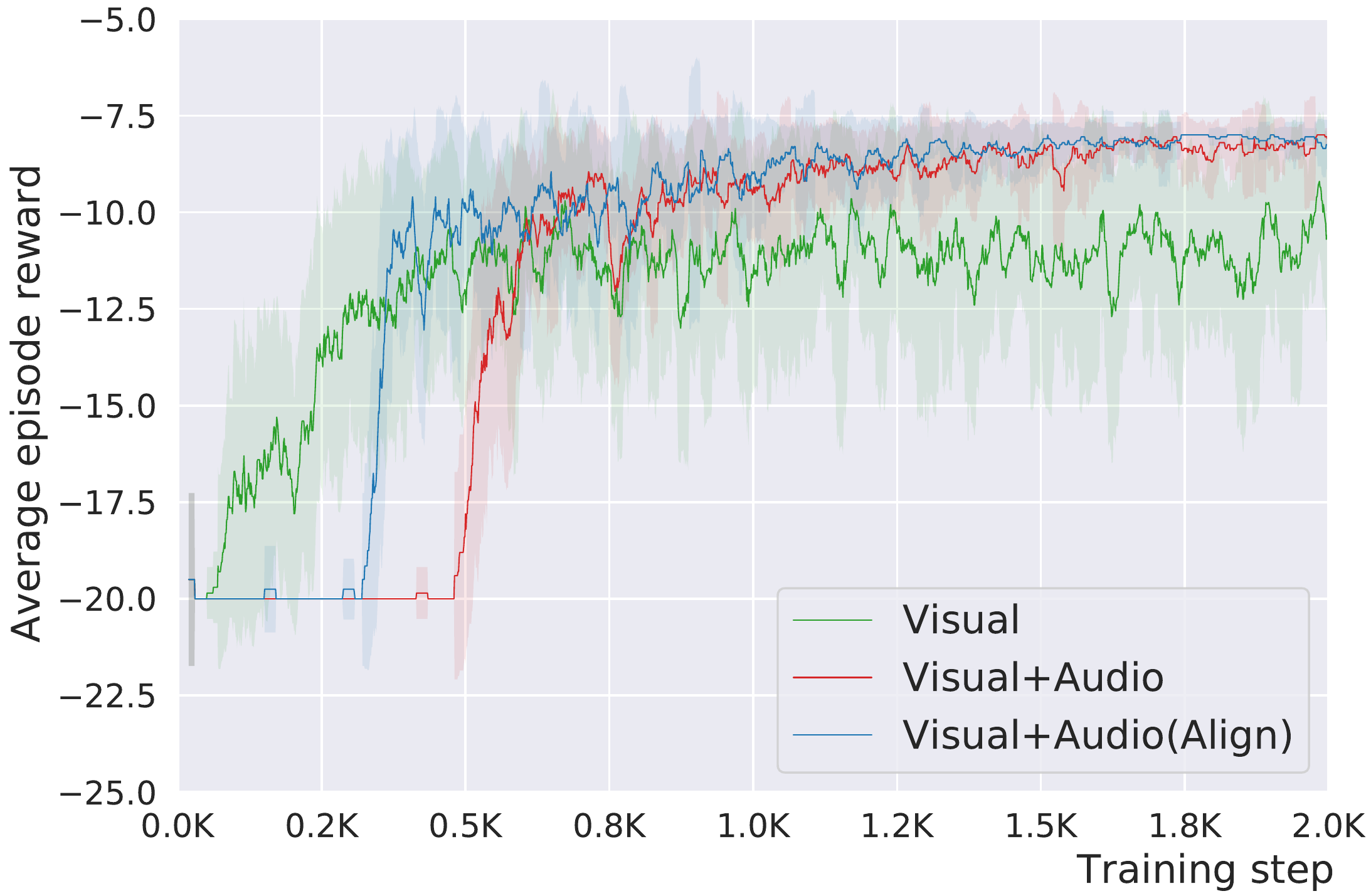}}
		\caption{Example for the impact of modal heterogeneity on RL (Given a sound source (black), the agent (white) perceives and relates it with the visual environment to navigate towards the source.).}
		\label{challenge_train}
	\end{figure}
	
	\subsubsection{Modal Heterogeneity}
	
	Intuitively, an agent with multimodal information can have a better understanding of the state. Indeed, this is true only when the information from different modalities is properly aligned. Otherwise, the agent may get ``confused'' if what it ``saw'' and what it ``heard'' are inconsistent. This is usually the case for a robot with multiple sensors. In order to test and verify the modal heterogeneity, we designed the gridworld domain as shown in Figure \ref{challenge_train} (a). For the agent (white block), it starts from the left-top corner and should reach the right-bottom corner (black block) by utilizing two modalities: visual and audio. The agent can complete the task through visual or audio input, i.e., the visual input provides the location of the elements and the audio input provides the direction of the sound source. However, if we use both modalities as input, it will weaken the ability of the agent to learn the policy due to the modal heterogeneity.
	
	As demonstrated by the simple example shown in Figure~\ref{challenge_train} (b), the agent with merely visual input (i.e., green line) can learn a policy surprisingly faster than the one combining with visual and audio data (i.e., red line). This result illustrates unaligned multi-modal information may cause "confusion" of the agent about the state. In other words, it is difficult to find a mapping directly from multimodal signals to actions from the indirect signal (reward function). Although an agent with multimodal information can eventually learn a better policy, it will require significantly more training data or may be difficult to converge in complex domains. As shown here, our approach with modality alignment (i.e., blue line), as described later, can speed up the training process while making use of the multimodal information to learn a better policy. Thus, we argue that modality alignment is important for the training of RL.
	
	
	\begin{figure}[t]
		\centering
		\subfigure[Gridworld domain.]
		{\includegraphics[width=0.38\hsize]{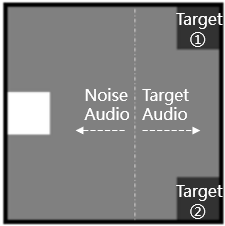}} 
		\subfigure[Training curves.]
		{\includegraphics[width=0.6\hsize]{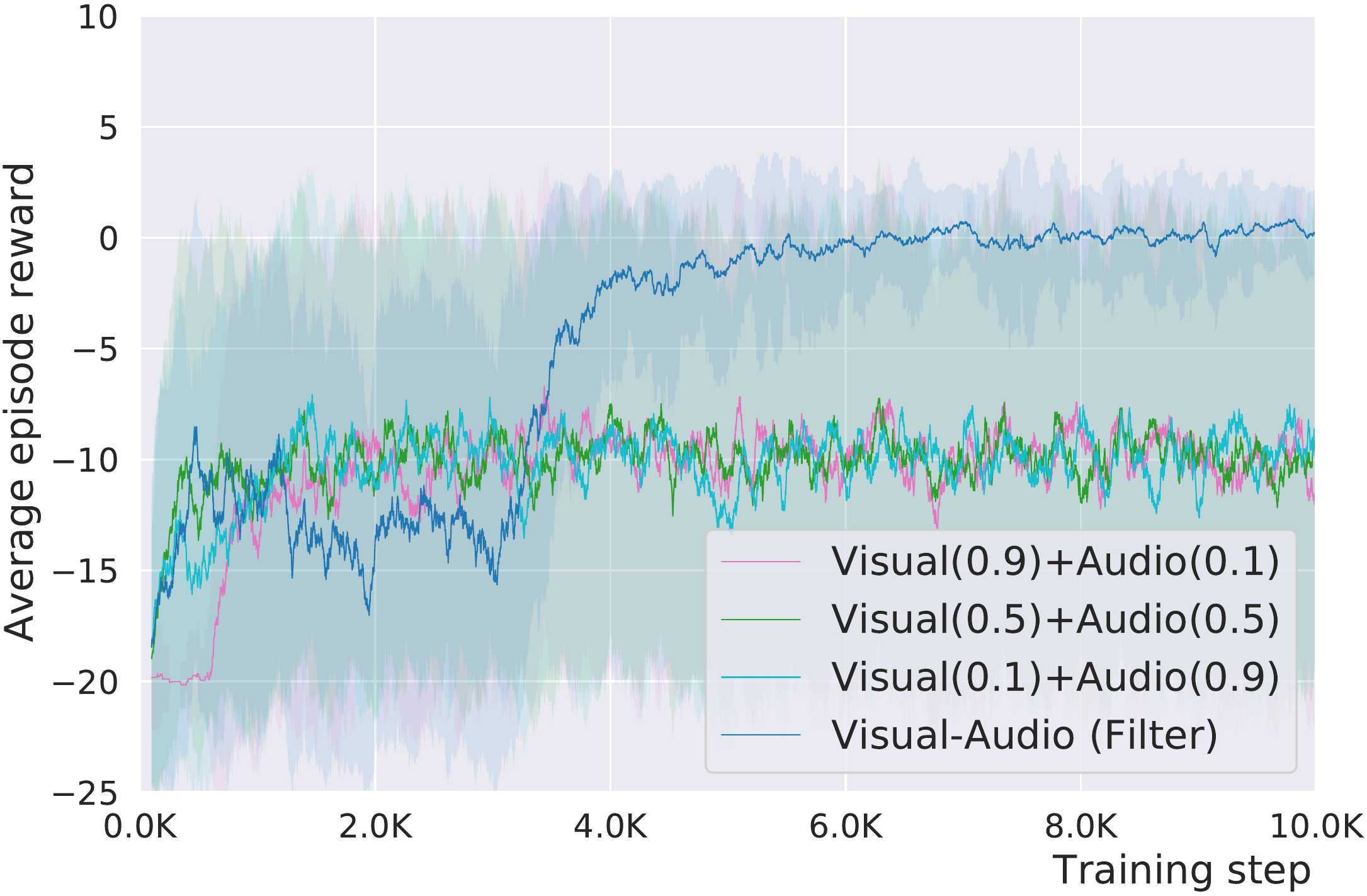}}
		\caption{Example for the impact of dynamic importance on RL (The task is for the white agent to reach the correct black location with both visual and audio information. Visual provides the position of different objects and audio identifies the type of target).}
		\label{challenge_train1}
	\end{figure}
	
	\subsubsection{Dynamic Importance}
	
	Information from different modalities can have different importance for decision making in different situations. Moreover, modality importance is dynamic and can shift from time to time. For example, people mainly rely on vision for driving but horns blaring may be more important for knowing another car around the corner. 
	
	Specifically, we designed the gridworld domain as shown in Figure \ref{challenge_train1} (a). For the agent (white block), it should reach target 1 (right-bottom corner) or target 2 (right-top corner) by making use of two modalities: visual and audio. The target type is randomly generated, and the audio indicates the associated target. The visual input is the raw image and provide the position of the elements, Figure\ref{challenge_train1} (a), but it can not distinguish which target should reach. So the audio provides the type of target where the agent should reach. When the agent's position is right on the white line, it will hear the audio information, distinguish which target should reach, and hears noise otherwise. In this case, it is critical for an RL agent to learn how to bias towards more valuable modalities at different moments. As shown in Figure~\ref{challenge_train1}, the agent with fixed modality importance failed to learn a good policy (e.g., the weights of visual and audio are [0.9, 0.5, 0.1] and [0.1, 0.5, 0.9] respectively). In contrast, our approach with the importance enhancement (e.g., blue line) can dynamically bias a certain modality to achieve much better performance. Therefore, the dynamic importance of modalities is important due to the sequential decision-making in RL.
	
	
	In summary, learning an effective state representation is particularly challenging for multimodal RL due to modal heterogeneity and dynamic importance. Although there is a rich literature on multimodal ML for modality alignment \cite{kim2016multimodal, vo2019composing, kahou2016emonets, sun2019learning} or information fusion \cite{morvant2014majority, chambers2014robust, nobili2017heterogeneous, huang2020multi}, most of them are devised for supervised learning with labeled data. In general, due to the sequential decision-making nature of RL, the training process of trial-and-error exploration, and the dynamic importance of modalities in the time scale, these methods cannot be directly applied to multimodal RL. Both challenges require specific treatments in the RL context as detailed next.

	
	\section{The Method}
	
	\noindent Here, we propose our multimodal RL (named MAIE, which stands for Modality Alignment and Importance Enhancement) with effective state representation learning. As aforementioned, state representation is challenging in multimodal RL due to modal heterogeneity and dynamic importance of different modalities. Respectively, we devise the modality alignment and importance enhancement modules to address these issues. We put them together to learn an effective state representation that can be used by deep RL methods (e.g., A2C). Specifically, we introduce a novel technique to produce a better state vector that can be used by the value or policy network training in an end-to-end manner with multimodal information.
	
	In more detail, our state representation networks are trained with two-step backpropagation in each episode: one is from the loss of state representation learning, and the other is from the loss of standard RL with a feedback of the reward signals. The former aims to updating the parameters of feature extractors and aligning the modalities in the embedding space. Then the latter is the normal backpropagation to update the parameters based on the RL procedure. Notice that this backpropagation also comes through the importance enhancement module so that networks of different modalities are updated with different importance weights. By doing so, our method is able to learn state representation that plays well with the RL process.
	
	In what follows, we will describe our modality alignment and importance enhancement in detail.
	
	\begin{figure}[t]
		\centering
		\includegraphics[width=1.0 \hsize]{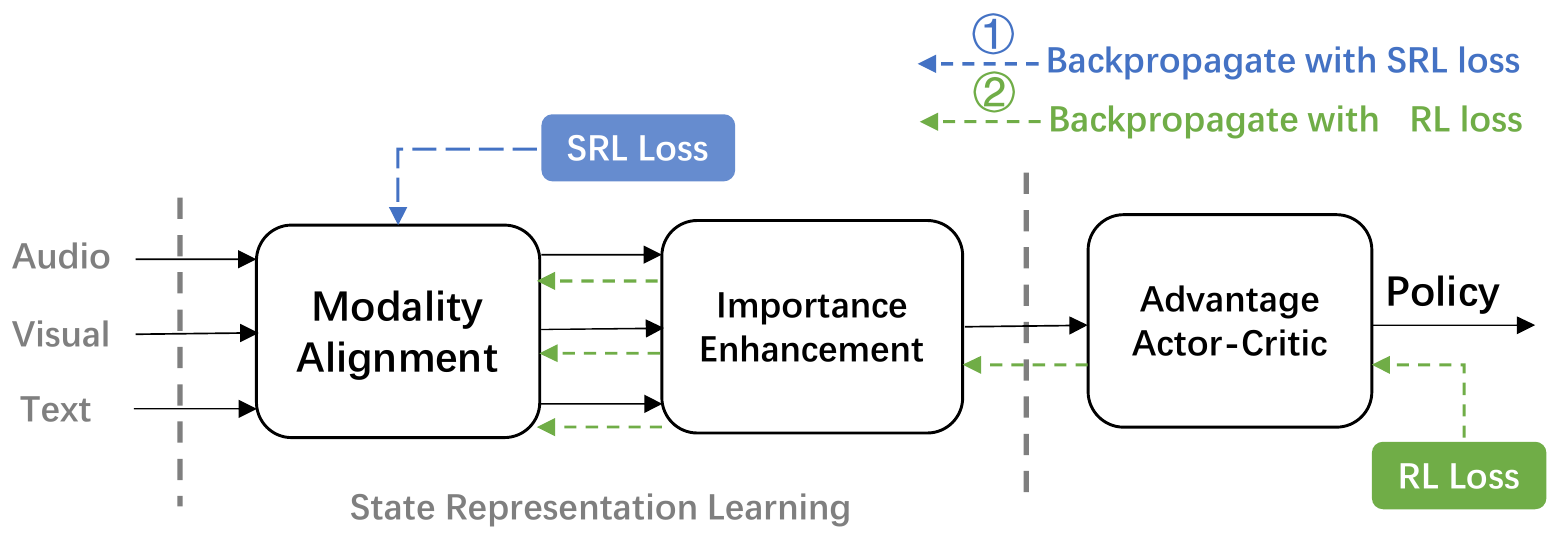} 
		\caption{Overall framework of MAIE with end-to-end training.}
		\label{Network1}
	\end{figure}
	
	\subsection{Modality Alignment}
	
\begin{figure}[t]
		\centering
		\includegraphics[width=1.0 \hsize]{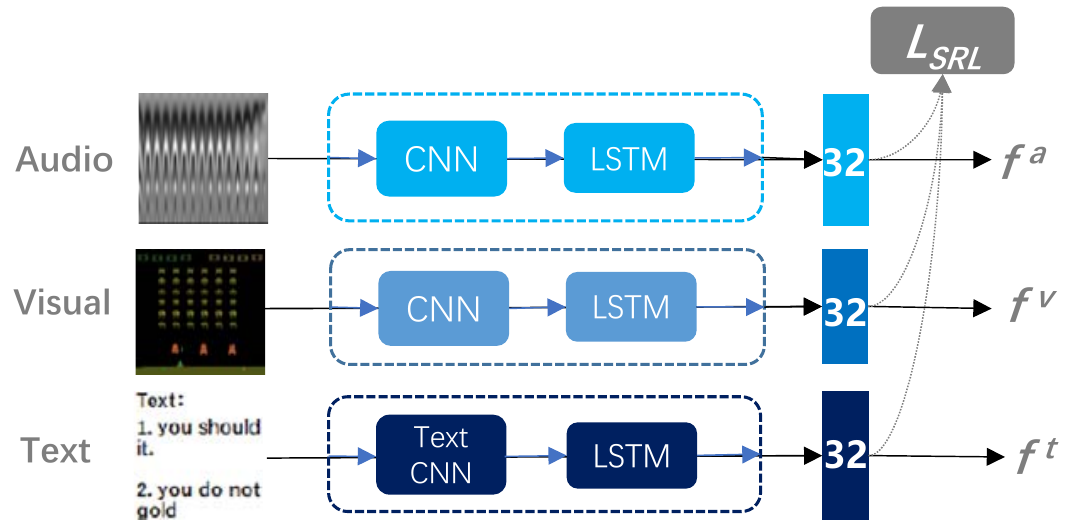} 
		\caption{Network architectures of modality alignment.}
		\label{Network2}
	\end{figure}

	\noindent As shown in Figure \ref{Network2}, we use CNN and LSTM to extract high-dimensional and temporal features of each modality. Without loss of generality, we consider three types of modalities (i.e., visual, audio, and text) that are commonly used in practice. Note that such feature extractors (i.e., CNN and LSTM) are widely used to train single-modality RL agent in an end-to-end fashion. They are useful to convert the raw data into vector embedding state features. 
	
	Here, we train a feature extractor $\mathcal{F}$ separately for every modality (e.g. image, audio, and text). To retrieve useful information from the raw data of image and audio, we first use CNN to extract features of each input. For text, we use TextCNN \cite{kim2014convolutional} to extract features, in which multiple kernels of different sizes are used to extract key information in sentences (similar to n-grams with multiple window sizes), to capture local correlations better. Then the features fed to the LSTM cell for capturing temporal information from the sequence. The final hidden state of LSTM is used as an input of the following network module, where $f^i$ is the image feature embedded by sensory feature extractor $\mathcal{F}^i$, and $f^a$, $f^t$ is the audio and text embedded feature. 
	
	As mentioned earlier, due to modal heterogeneity, simply concatenating the vectors of all modalities may not be helpful for the RL training. Therefore, we introduce two module, i.e., {\em similarity aggregation} and {\em temporal discrimination}, in order to train extractor of different modalities to generate vectors that are properly aligned in the embedding space.
	
	
	\subsubsection{Similarity Aggregation} 
	
	Due to modal heterogeneity, different modalities are usually distributed inconsistently by the feature extractors. However, in most cases, there is a synchronous nature in these modalities, meaning that the states expressed by different modalities can be consistent at the same time. For example, when people speak, we can hear the voice and see the lip movements. Intuitively, it would be helpful if we can find relationships and correspondences between sub-components of instances from two or more heterogeneous modalities. We denote the similarity aggregation as follow.
	
	\newtheorem{definition}{Definition}
	\begin{definition}[Similarity aggregation]
		Let $F_x(X) = [x_1, x_2, \dots]$ and $F_y(Y) = [y_1, y_2, \dots]$ denote features of two modalities. The similarity aggregation aims to align feature $x_i$ with $y_j$, both of which correspond to the same state attribute $s_k$, where the true state $S$ has attributes: $S = [s_1, s_2, \dots]$.
		\label{define:similarity aggregation}
	\end{definition}
	
	To achieve this, we use a similarity measurement (e.g., Euclidean, cosine, and KL distance), which are commonly used in multimodal machine learning \cite{aytar2017see}, to do multimodal alignment through a loss.
	Specifically, let $f^i$ and $f^j$ be two vector representations of modalities $i$ and $j$ that need to be aligned in the feature space. Given this, we define the loss function as follow:
	\begin{equation}
		\label{Loss-similar}
		\mathcal{L}_{sim}(\phi) = \sum_{i=1}^{m} \sum_{j \neq i} \psi [f^i \mid f^j]
	\end{equation}
	where $\psi [f^i \mid f^j]$ is a distance function for vectors $f^i$ and $f^j$.
	
	
	
	
	\subsubsection{Temporal Discrimination} 
	
	Multimodal alignment based on similarity is particularly useful when all the modalities are informative and corresponding to the underlying state. However, when one of the modalities is totally not informative or even purely noisy (e.g., visual modality in a completely dark room), forcing other modalities to align with it will reduce their temporal discrimination and make them less and less sensitive.
	This issue is especially critical for sequential decision-making problems. Therefore, we expect that the feature vectors of identical modalities are discriminative in the time scale. We denote the temporal discrimination as follow.
	
	\begin{definition}[Temporal discrimination]
		Let $X$, $Y$ denote two modalities. Assume that $X$ remains unchanged during consecutive timesteps, e.g., $X_t = X_{t+1}$. After modality alignment between $X$ and $Y$ at each timestep, the feature of $Y_t$ and $Y_{t+1}$ will be smoothed and tend to be less distinctive. However, we expect that $Y_t$ and $Y_{t+1}$ are discriminative.
		\label{define:temporal discrimination}
	\end{definition}
	
	To this end, temporal discrimination is introduced for increasing the difference between the feature vectors of each modal at different moments with the loss function defined as:	
	\begin{equation}
		\label{temporal difference}
		\mathcal{L}_{td}(\phi) = - \sum_{i=1}^{m} \sum_{t=1}^{T-1} \psi [f_{t}^i \mid f_{t+1}^i]
	\end{equation}
	
	
	To put the loss functions in Equations \ref{Loss-similar} and \ref{temporal difference} together, we have the overall loss function for the backpropagation of state representation learning:
	\begin{equation}
		\label{srl_loss}
		\mathcal{L}_{SRL}(\phi) = c_{sim} \mathcal{L}_{sim}(\phi) + c_{td} \mathcal{L}_{td}(\phi)
	\end{equation}
	where $c_{sim}, c_{td}$ are scaling constants. 
	
	
	Similar to multimodal ML \cite{baltruvsaitis2018multimodal}, our purpose of the {\em modality alignment} module is to find relationships and correspondences between sub-components of instances from multiple heterogeneous modalities. Note that RL is a sequential decision-making problem. Therefore, we integrated a {\em temporal discrimination} to discriminate the feature vectors of identical modalities in the time scale. As shown earlier in Figure \ref{challenge_train}, our modality alignment module indeed is able to address the modal heterogeneity issue and speed up the RL training. This is also confirmed later by our experiments.
	
	\begin{figure}[t]
		\centering
		\includegraphics[width=1.0 \hsize]{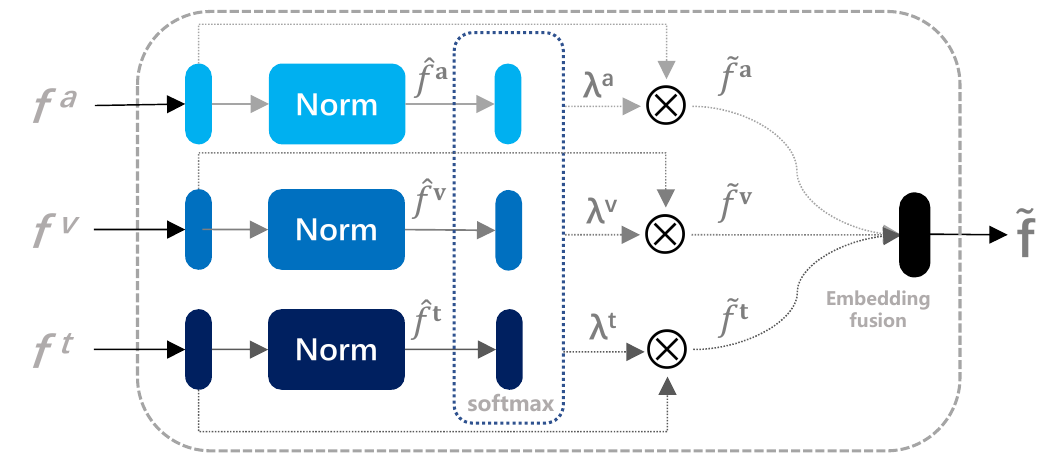} 
		\caption{Network architectures of importance enhancement.}
		\label{Network3}
	\end{figure}
	
	\subsection{Importance Enhancement}
	
	\noindent Now, we proceed to our importance enhancement module, which is inspired by work \cite{ioffe2015batch}. As aforementioned, our purpose of using the importance enhancement is to bias towards more crucial modality when the importance of modalities changes. As shown in Figure \ref{Network3}, this is done by putting embedded feature vectors into an enhancement network and computing the importance coefficient.
	
	Specifically, we first normalize feature vector $f^m$ of each modality $m$ as follow:
	\begin{equation}
		\label{norm}
		\hat{f}^m = \frac{f^m - \mu^m} {\sqrt{\sigma^m + \epsilon}}
	\end{equation}
	where $\mu^m, \sigma^m$ are the mean and variance computed by soft-update as in Equations \ref{mean} and \ref{var}, and constant $\epsilon$ is added to the variance for numerical stability. Noting that both the mean and variance are regarded as additional parameters saved in the network when training and retrieved during inference.
	
	
	Generally, it is impractical to computed based on the entire training set. Therefore, we use mini-batches and soft-update to estimate the mean and variance. Specifically, we consider a mini-batch $\mathcal{B}$ of size $k$, $\mathcal{B} = \{\tau_1, \tau_2,\cdots, \tau_k\}$. For each modality, we calculate and update the mean and variance of each modality $m$ as follow:
	
	\begin{equation}
		\label{mean}
		\mu_{\mathcal{B}}^m = \frac{1}{|\mathcal{B}|} \sum_{i=1}^{|\mathcal{B}|} f_i^m
	\end{equation}
	
	\begin{equation}
		\label{var}
		\sigma_{\mathcal{B}}^m = \frac{1}{|\mathcal{B}|} \sum_{i=1}^{|\mathcal{B}|} ( f_i^m - \mu_{\mathcal{B}}^m )^2
	\end{equation}
	
	Then, the mean and variance of modality $m$ are updated with the decay factor $\xi$ as follow: 
	
	\begin{equation}
		\label{moving}
		\mu^m = \xi \cdot \mu_{\mathcal{B}}^m + (1 - \xi) \cdot \mu^m_- ~,~~~~~~~ 
		\sigma^m = \xi \cdot \sigma_{\mathcal{B}}^m + (1 - \xi) \cdot \sigma^m_- 
	\end{equation}

	Here, we consider the normalized features $\hat{f}^m$ is more informative if the feature deviates further from its mean since it occurs with a lower probability and often provides more information \cite{informationTheory}. For example, when people drive on road, unexpected loud noise usually means something about the car went wrong and should be alerted. Based on this property, we use softmax to calculate the importance coefficient $\lambda^m$ for each modality $m$ as follow:
	\begin{equation}
		\label{softmax}
		\lambda_{[l]}^m = \frac{e^{|\hat{f}_{[l]}^m|}} {\sum_{i=1}^{|\mathcal{M}|} e^{|\hat{f}_{[l]}^i|}}
	\end{equation}
	where $\hat{f}_{[l]}^m$ is $l$-dimensional of feature $\hat{f}^m$, $|\mathcal{M}|$ denotes the set of all modalities that agent can access. 
	
	
	During forward inference, we perform the element-wise product of $\lambda^m$ and $f^m$ to get the weighted features: $\tilde{f}^m = \lambda^m \cdot f^m$. Then, we concatenate the weighted features, $\tilde{f} = [\tilde{f^1}, \tilde{f^2},..., \tilde{f^m}]$, as a state representation for the standard RL.

	At training time, the gradient of RL loss will backpropagate through this importance enhancement. As a result, the gradient is computed with respect to the importance coefficient of multiple modalities. Given modality $m$ with the input $f^m$ and output $\tilde{f}$, by using the chain rule, we have the following equations:
	\begin{displaymath}\small
		\begin{aligned}
			\frac{\partial \mathcal{L}}{\partial f^m} &= \frac{\partial \mathcal{L}}{\partial \tilde{f}} \cdot \frac{\partial \tilde{f}}{\partial \tilde{f^m}} \cdot \frac{\partial \tilde{f^m}}{\partial f^m}  = \frac{\partial \mathcal{L}}{\partial \tilde{f}} \cdot \frac{\partial \lambda^m \cdot f^m}{\partial f^m} \propto \frac{\partial \mathcal{L}}{\partial \tilde{f}} \cdot \lambda^m
		\end{aligned}
	\end{displaymath}
	where $\mathcal{L}$ is the gradient of the actor-critic method as computed by Equations \ref{Loss-critic} and \ref{Loss-actor}. 
	
	
	Notice that the features and gradients of different modalities are weighted by the importance coefficient $\lambda^m$ respectively in the forward and backpropagation phases. Meanwhile, the gradient of RL loss also depends on the rewards obtained by the agent interacting with the environment. By doing so, the crucial modality will be amplified to play a more important role in the decision-making, and it also ensures the agent learn a correct policy based on key observations. This is similar to the attention mechanism \cite{omidshafiei2017crossmodal}. However, the attention mechanism may require a large number of trajectories to fit the network. Alternatively, we offer an easier-to-train and more explainable approach for multimodal RL. As shown in Figure \ref{challenge_train1} and confirmed later in our experiments, our method can significantly improve the policy quality especially in domains where the modality importance varies in the process.
	
	\section{Experiments}
	\noindent In real-world multimodal applications, there are generally two typical settings: 1) redundant modalities, where every modality provides the complete state information; and 2) complementary modalities, where every modality only provides partial information about the state. Both settings require different modalities to be properly aligned. In particular, the second setting is more difficult because the agent may need to bias towards certain modalities accordingly. For evaluating the effectiveness of our approach, we developed three benchmark domains: Audio-Visual Navigation, Mining and autonomous driving. In Audio-Visual Navigation, the agent can navigate to the goal location using either visual or audio information (i.e., redundant modalities). The Mining domain is more challenging because the agent needs to make full use of multiple modalities and the modality importance will change during the task (i.e., complementary modalities). Finally, we performed a case study in a more challenging and realistic domain: self-driving car control, which aims to show the potential and usefulness of our approach in real-world RL applications. Therefore, by testing on these domains, the performance of our modality alignment and importance enhancement modules can be well evaluated.
	
	\begin{figure}[t]
		\centering
		\subfigure[]
		{\includegraphics[width=0.24\hsize]{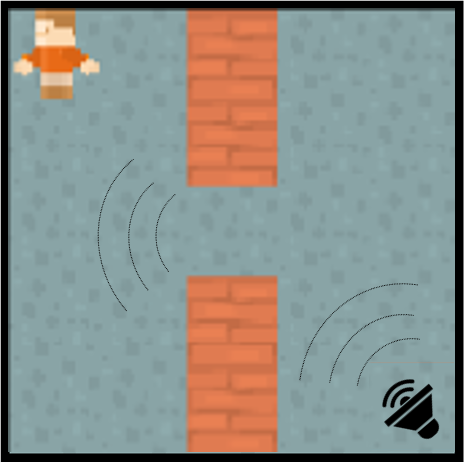}}
		\subfigure[]
		{\includegraphics[width=0.24\hsize]{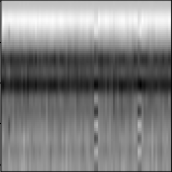}}
		\subfigure[]
		{\includegraphics[width=0.24\hsize]{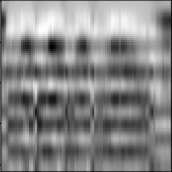}}
		\subfigure[]
		{\includegraphics[width=0.24\hsize]{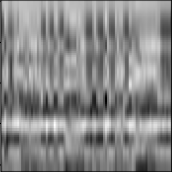}}
		\caption{Visual and audio modalities in Audio-Visual Navigation. (a) represents the visual input of map image. (b-d) represent the stereo audio channels, right audio channel and left audio channel respectively.}
		\label{Navigation}
	\end{figure}
	
	\begin{figure}[t]
		\centering
		\subfigure[Training curves (Audio-Visual Navigation)] {\includegraphics[width=0.85\columnwidth]{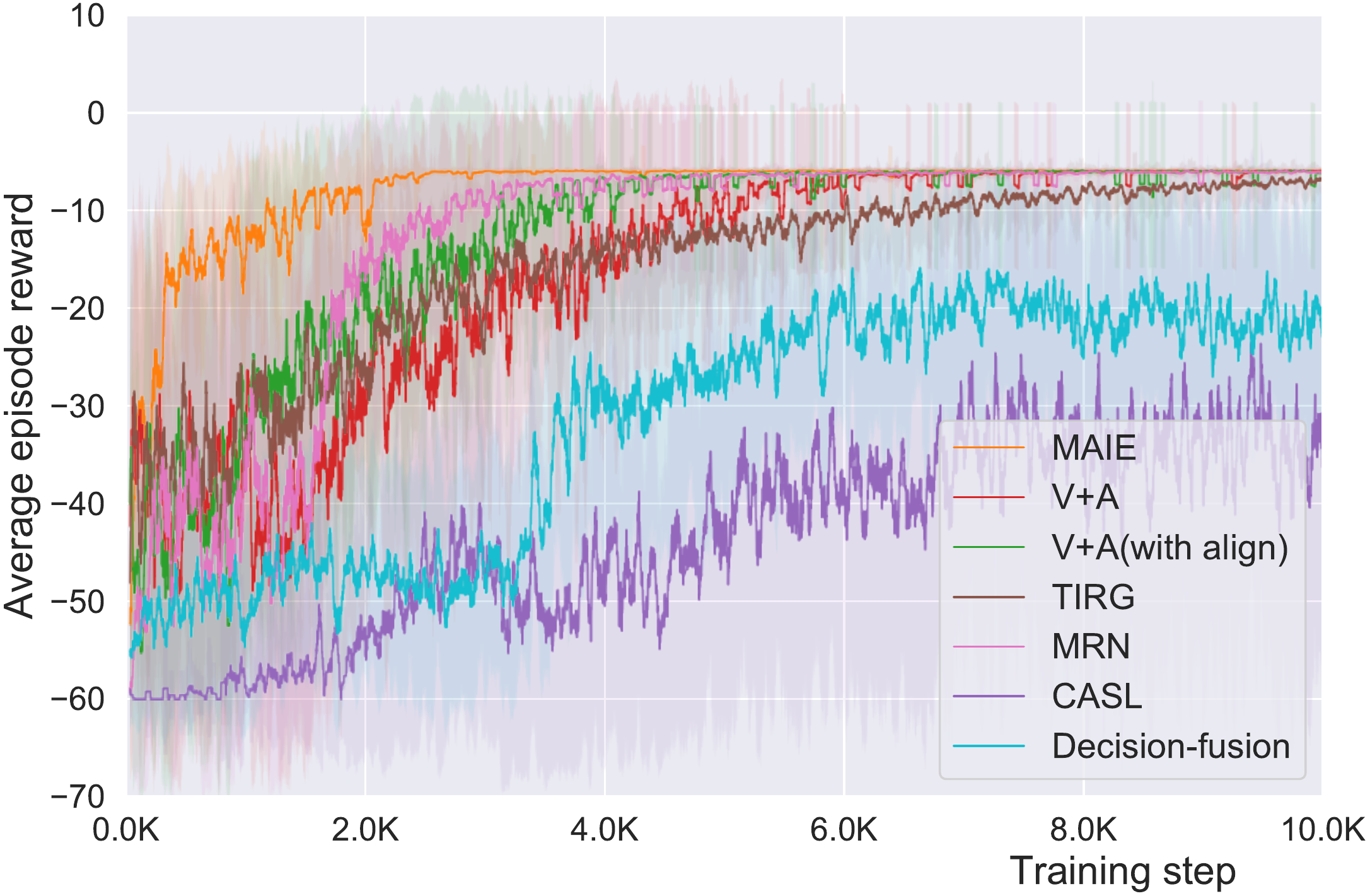}}
		\subfigure[Training curves (Mining)] {\includegraphics[width=0.85\columnwidth]{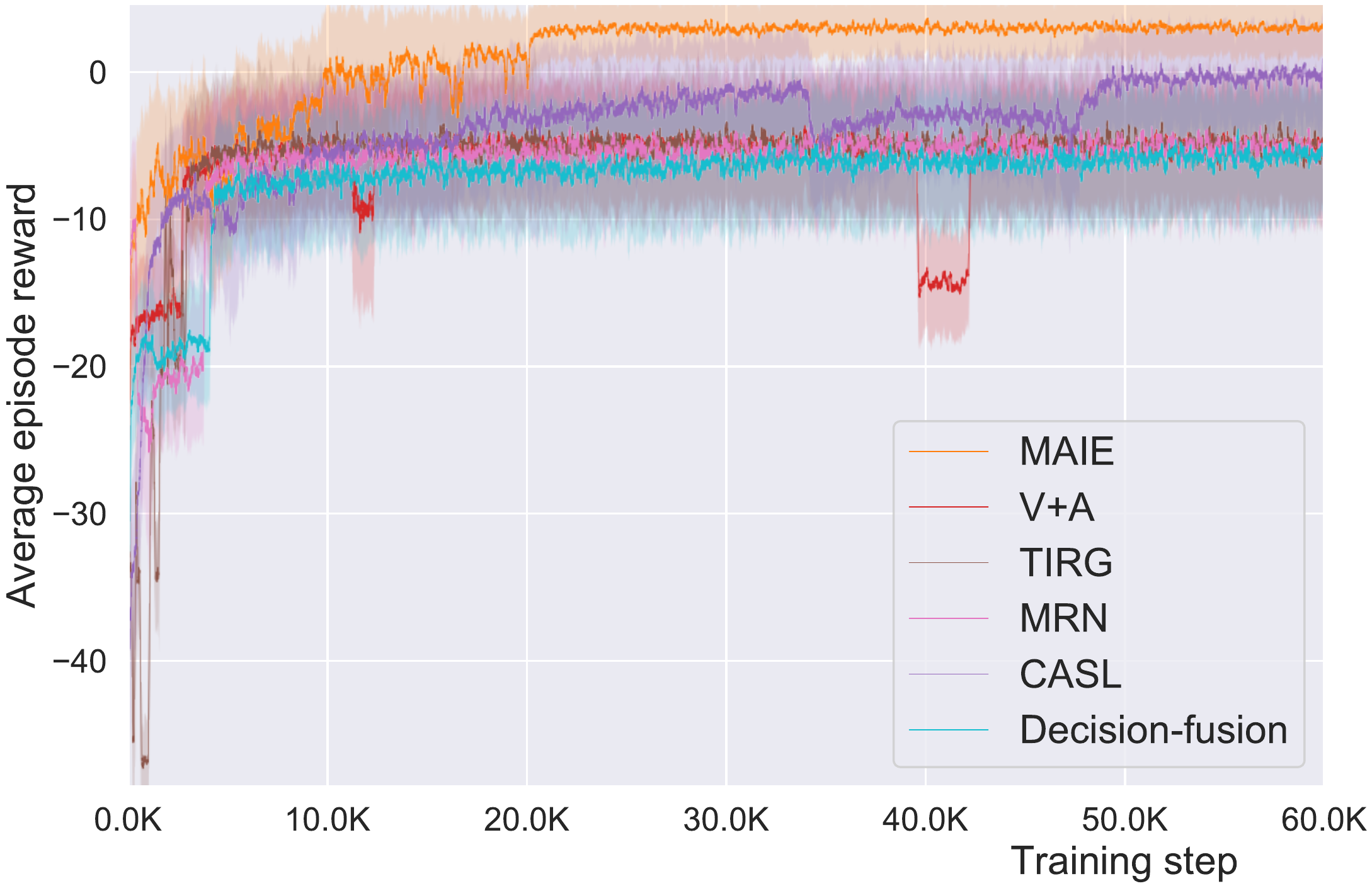}}
		\subfigure[Training curves (Mining+)] {\includegraphics[width=0.85\columnwidth]{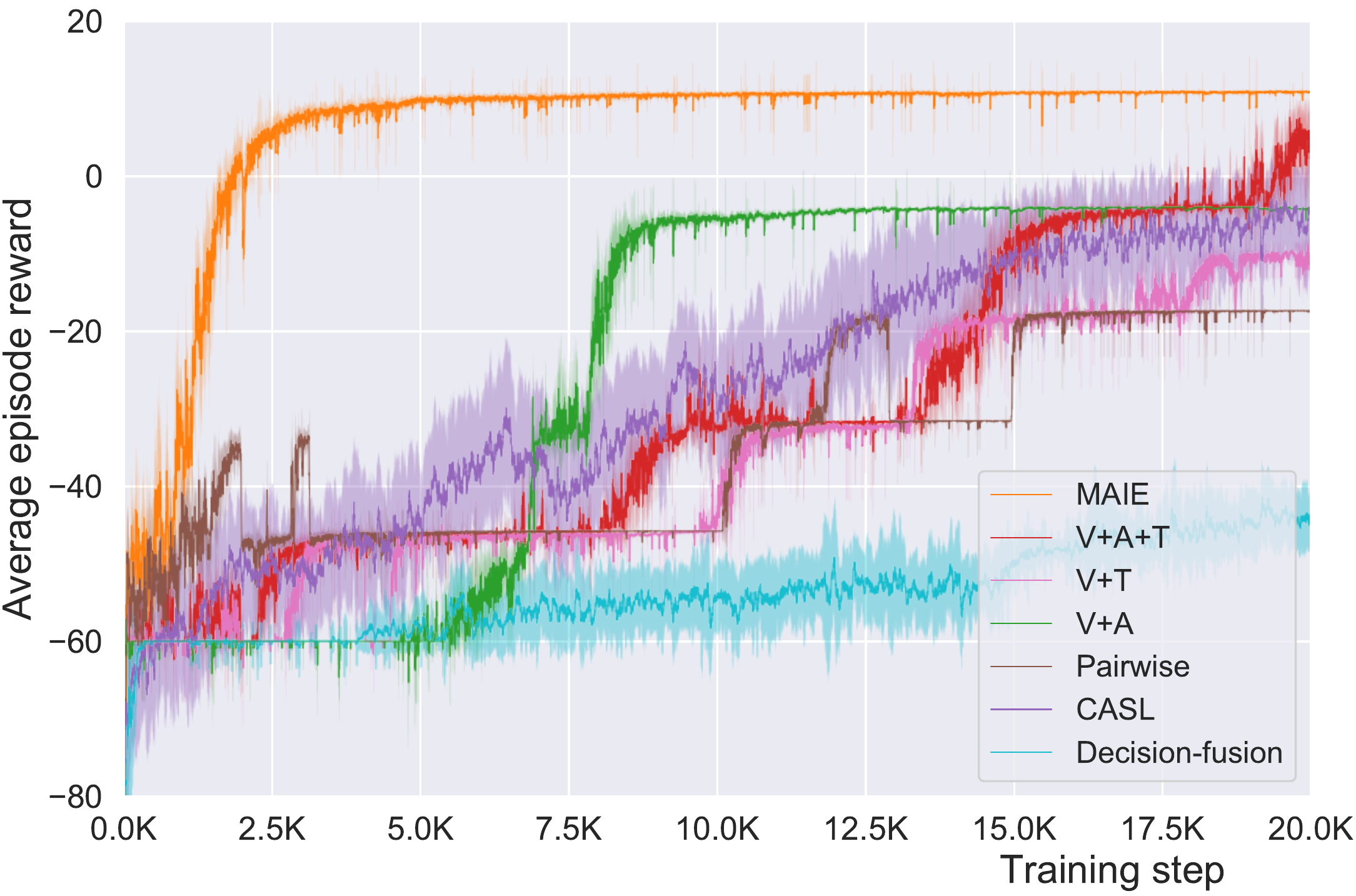}}
		\caption{Training curves for the three domains (the solid line and shade are the mean value and standard deviation respectively), plotted based on three training runs using different random seeds.}
		\label{Training-curves}
	\end{figure}
	
	\subsection{Network architectures}
	In the experiment, we used the following neural network architecture. Two separate feature extraction were employed:
	
	\begin{itemize}
		\item For the visual and audio modalities, the visual and audio features are extracted by the feature extractor composed of CNN and LSTM. The architecture of CNN contains three convolutional layers with 32 filters. The filters are parameterized by kernel sizes 3, strides 2, and paddings 1. ReLU activations are used throughout the extractors. 
		\item  For the text modality,  the text features are extracted by the feature extractor composed of TextCNN and LSTM. The architecture of TextCNN  ~\cite{kim2014convolutional} contains three convolutional layers with 3 filters, parameterized by kernel sizes 2, strides 1, paddings 1 and ReLU activations.
	\end{itemize}
	
	\noindent All feature extraction of visual audio and text are followed by LSTM ~\cite{hochreiter1997long} to extract temporal information, which contains 32 hidden units. For the network of Actor-Critic, we both use two layers of fully connected with 256 and 64 units, respectively. Optimization was done with the ADAM, using a learning rate of 1 · 10$^{- 4}$.

	
	\subsection{Audio-Visual Navigation}
	
	\noindent In this domain, as shown in Figure \ref{Navigation}(a), an agent in a grid world must reach the target location as soon as possible. It can navigate to the goal with either a map image visually showing its current and target locations, or a sound that can vocally direct it towards the source (i.e. the target location). The setting of the audio is that when the audio source is horizontal with the agent, the stereo channel will be heard; when the sound source is above or below the agent, the left channel and the right channel will be heard respectively. To make it more challenging, walls are placed in the middle and the sound can only send through the corridor. Note that the agent can obtain audio information anywhere in the rooms. In the left room, the sound can only transmit through the corridor (i.e., the sound source of the left room) while in the right room the sound comes from the target. The agent will receive +1 reward when he completes this task and -1 for one-step cost.
	In the domain, the visual and audio modalities are redundant here so that the agent can reach the goal solely on any modality, i.e., downward when hearing the right channel, rightward when hearing stereo channel.
	The agent can complete the task through one of visual and audio, but it would be helpful and more robust if the agent can find relationships and correspondences between visual and audio modalities.
	
	We compared our approach \textbf{MAIE} with the following methods: 1) \textbf{V+A}: The simple baseline to directly concatenate the visual and audio inputs, used in multimodal RL \cite{henkel2019score}; 2) \textbf{V+A (with align)}: The V+A baseline enhanced by our modality alignment module;
	3) \textbf{Decision-fusion}: The method based on decision-level fusion for multimodal learning \cite{morvant2014majority}; 4) \textbf{TIRG}: The method based on study feature composition for multimodal learning \cite{vo2019composing}; 5) \textbf{MRN}: The method based on study joint representation from multimodal information for multimodal learning \cite{kim2016multimodal}; 6) \textbf{CASL}: The leading method based on attentive mechanisms for multimodal RL \cite{omidshafiei2017crossmodal}. Note that 1), 6) represent the state-of-the-art methods of multimodal RL and 3), 4), 5) are the common fusion techniques used in the general multimodal learning literature. For analytical purposes, 2) was also included. 
	
	As shown in Figure \ref{Training-curves}(a), our method converged much faster and more stably than other tested methods. \textbf{CASL} did not perform well because the attentive mechanism is hard to train and not crucial for this domain. In \textbf{Decision-fusion}, we observed the phenomenon of constantly switching modalities especially in the early stage slowed down the convergence. We can observed that \textbf{TIRG} and \textbf{MRN} achieve good performance thanks to learn the relationship between modalities. The improvement made by \textbf{V+A (with align)} comparing to \textbf{V+A} confirms the effectiveness of our modality alignment module. Our importance enhancement module is also helpful because the audio provides more direct guidance than the visual in this problem. Therefore, our method has the overall best performance in this domain.
	
	\subsection{Mining Domain}
	
\begin{figure}[t]
		\centering
		\subfigure[] {\includegraphics[width=0.24\hsize]{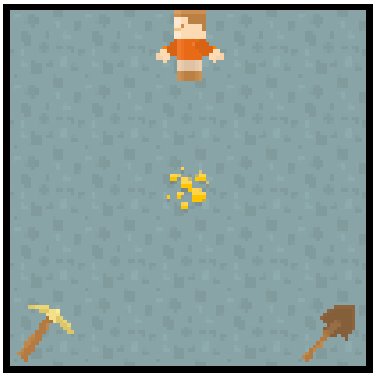}}
		\subfigure[] {\includegraphics[width=0.24\hsize]{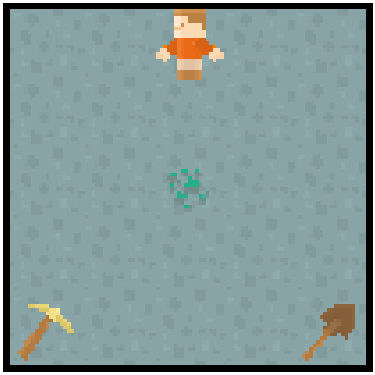}}
		\subfigure[] {\includegraphics[width=0.24\hsize]{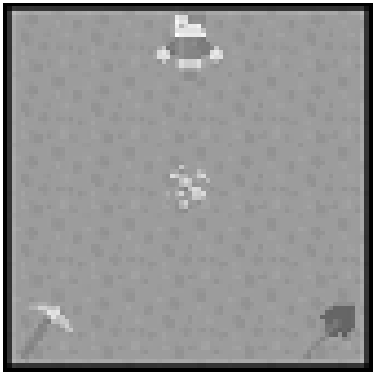}}
		\subfigure[] {\includegraphics[width=0.24\hsize]{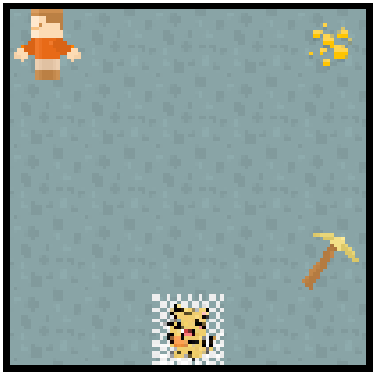}}
		\caption{Mining domain. (a-b) represent different ores require different tools, and (c) represents that ore type indistinguishable by the gray visual input. (d) represent the Mining+ with monster.}
		\label{MineCraft}
	\end{figure}

	\noindent This domain is originally introduced by the \textbf{CASL} paper \cite{omidshafiei2017crossmodal} for testing multimodal RL. As shown in Figure \ref{MineCraft}(a-b), an agent wants to mine either gold or iron ore but it must pick an appropriate tool before that. Note that the gold or iron ores appear identically in the gray image (Figure \ref{MineCraft}(c)), which is the visual input of the agent. Hence it must determine the type of ores based on their unique audio cue and then pick the right tool for mining. Here, visual input is useful for navigation, and the audio is necessary when deciding which tool to pick.
	
	To make it more challenging, we extended the Mining domain with more modalities (i.e., visual and audio plus text), namely Mining+, as shown in Figure \ref{MineCraft}(d). Now, the agent's task is to pick the tool to mine the gold while avoiding the monster. When approaching gold, tool, or monster, the agent will get unique audio and corresponding text hint (Table \ref{tableT}), which can be used as high-level tips to complete the task. The agent receives +10 reward for reaching gold, -100 for hurt by the tiger, and -1 for step cost. 
	In order to succeed in both domains, the agent must properly align all the modalities and bias towards certain ones that are more important. 
	
	\begin{table}[t]
		\centering
		\caption{List of text messages in the Mining+.}
		\begin{tabular}{cc}
			\toprule
			Point & Text \\ 
			\midrule
			1  & You should find gold and mine gold.\\
			2  & You do not have ax.\\
			3  & You get the ax, go to mine gold.\\
			4  & You get gold.\\
			5  & You hurt by tiger.\\
			\bottomrule
		\end{tabular}
		\label{tableT}
	\end{table}
	
	As in the previous experiments, we compared our approach with several state-of-the-art methods in the multimodal RL literature. Similarly, \textbf{V+A+T, V+A, V+T, A+T} denote the baselines directly concatenating inputs of multiple modalities (i.e., \textbf{V}isual, \textbf{A}udio, and \textbf{T}ext). Additionally, we included \textbf{Pairwise}: the method with pairwise concatenation \cite{sun2019learning}. For example, given visual, audio, and text, the pairwise concatenation can be as: [(\textbf{V} + \textbf{T}) + (\textbf{V} + \textbf{A})]. Note that it is only applicable for the Mining+ domain because it requires at least three modalities.
	
	As shown in Figure \ref{Training-curves}(b-c), our method substantially outperformed all the compared methods, both in the speed of convergence, the stability of learning, and the quality of policy. Again, this confirms that our method can effectively align the modalities and dynamically enhancement them based on their importance. The reason why these state-of-the-art machine learning methods (i.e. \textbf{TIRG}, \textbf{MRN} and \textbf{Pairwise}) did not achieve good results in this environment is that modalities may play different importance at some periods in dynamic environments, but they can not handle this.
	
	The competitive performance of CASL shows that the attention mechanism is indeed useful in this domain. However, the training of the attention mechanism requires a large number of samples. The reason is that the regular attention mechanism require additional trainable parameters. These trainable parameters tend to the overfitting to the training dataset generated from the agent's exploration in the environment. Particularly, in the early stage, since the weights assigned to the new states are not adequately optimal and are a kind of randomness, the generated training dataset is more or less unreliable. This tends to mislead the regular attention method into fitting undesired relationships. 
	Different from the regular attention mechanism, our importance enhancement can perform self-adapting but also requires no additional trainable parameters. No additional trainable parameters avoid the potential overfitting to the training dataset. The importance enhancement is only instance-based, so it can be self-adaptive for a specific instance in the early exploration. It performs the enhancement of the special modality to play a more important role in decision-making. Therefore, CASL converged much slowly than ours. All in all, we advances the state-of-the-art with a more effective and efficient multimodal RL approach.
	
	\subsection{Ablation Experiments}
	
	\begin{figure}[t]
		\centering
		\includegraphics[width= 0.9\hsize]{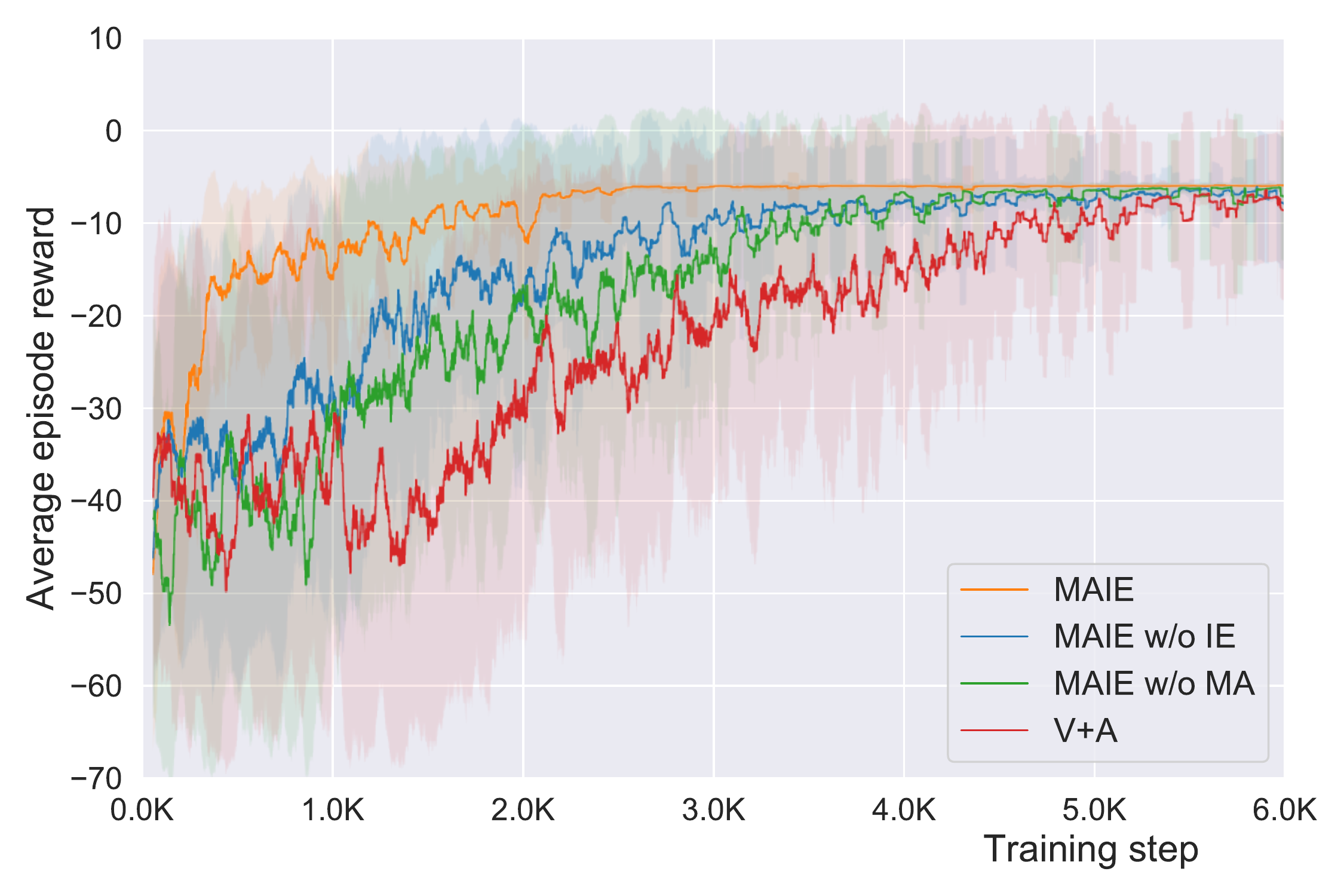}
		\caption{Ablation experiments for different modules.}
		\label{ablation}
	\end{figure}
	
	\noindent Here, we study the usefulness of our {\em modality alignment} and {\em importance enhancement} modules on the Audio-Visual Navigation scenarios. We consider training MAIE without {\em modality alignment}, MAIE without {\em importance enhancement} and MAIE without {\em modality alignment} and {\em importance enhancement} (i.e., MAIE w/o MA, MAIE w/o IE, and V+A). 
	
	As shown in Figure \ref{ablation}, the result shows that the importance enhancement improves the speed as the policy trained without it has a significant decrease in the performance. As aforementioned, the importance enhancement can be self-adaptive for a specific instance and generate the useful modality weights in the early exploration. 
	Meanwhile, the modality alignment improves the stationarity and speeds up the training process, which thanks to avoiding modal heterogeneity and aligning different modalities to have a better understanding of the state. In general, the modules in our method all play an importance role in multimodal RL.

	\subsection{Module Analysis}
	
	\noindent We conducted empirical analysis on our {\em modality alignment} and {\em importance enhancement} modules, to see how effective state representation can be learned in our method. As mentioned, this is key to the success of multimodal RL methods.
	
	\begin{figure}[t]
		\centering
		\includegraphics[width=1.0\columnwidth]{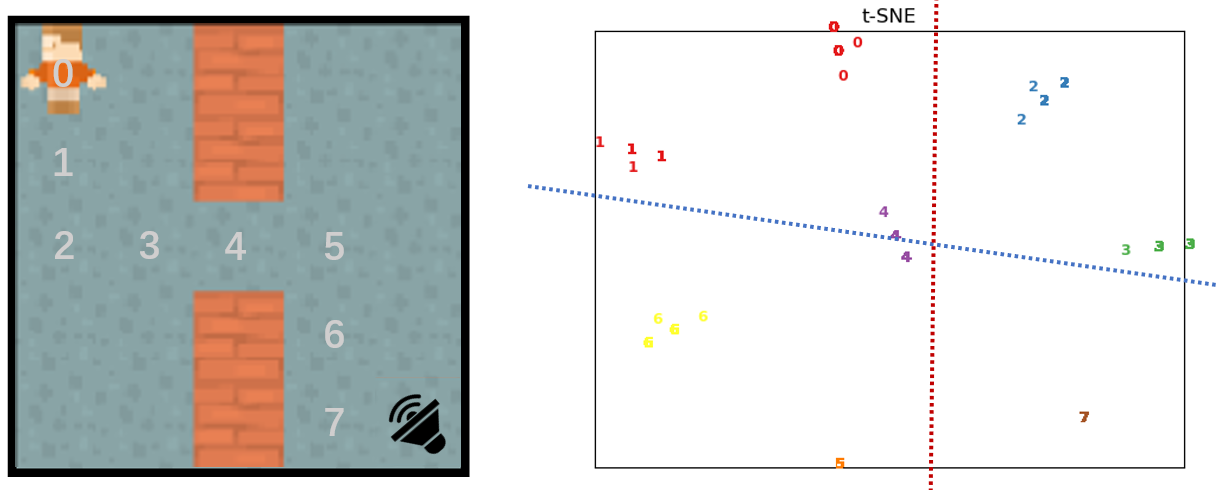}
		\caption{Results of the t-SNE method on modality alignment.}
		\label{fig:tsne}
	\end{figure}

	\subsubsection{Modality Alignment} 
	
	We use the t-SNE ~\cite{maaten2008visualizing} method to project the state of an episodic trajectory, learned by our method, into a 2-D plane. Figure \ref{fig:tsne} shows a division phenomenon of visual and audio modalities with the blue and red lines respectively. Take the trajectory with locations \{0, 1, ... , 7\} in Audio-Visual Navigation for instance. We can see that the locations \{0, 1, 2, 3\} on the left side of the wall is above the blue line, while the locations \{5, 6, 7\} on the right side of the wall are below. Similarly, the locations \{0, 1, 4, 5, 6\} with the right audio channel are on the left of the red line, which the locations \{2, 3, 7\} with the stereo audio channels are on the right. This shows that the visual and audio modalities are properly aligned (otherwise locations will be on the wrong side) with our module and effective state representations (i.e., locations for the navigation task) are learned. 
	
	\begin{figure}[t]
		\centering
		\includegraphics[width=1.0\columnwidth]{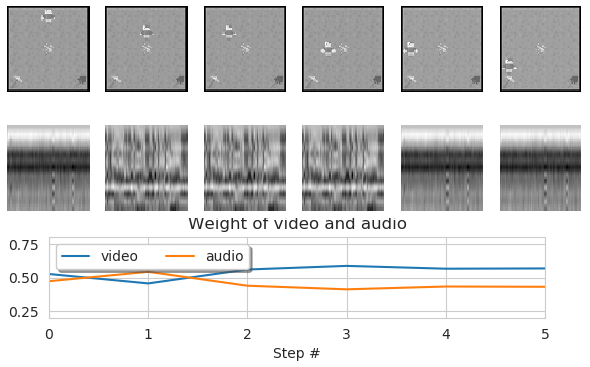}
		\caption{Results of modality weights on importance filter.}
		\label{fig:analysis}
	\end{figure}
	
	\begin{figure}[t]
		\centering
		\includegraphics[width=1.0\columnwidth]{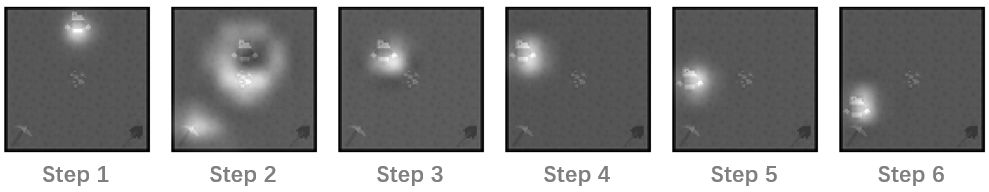}
		\caption{Illustrate of learning what features are important. We display the policy saliency in white, representing features the agent pays attention to.}
		\label{fig:analysis2}
	\end{figure}
	
	\subsubsection{Importance Enhancement} 
	
	We plot the importance coefficient $\lambda^m$ of each modality $m$ extracted from our \emph{ importance enhancement network} during an episode frames in the Mining domain. Note that the importance coefficient is used in our module to weight that modality and thus bias towards modalities with high coefficient values. As shown in Figure \ref{fig:analysis}, the weight of the audio modality increases at step 1, as the agent is currently near the ore and needs to know its type through the audio. After that, the audio weight decreases because the visual is more important for navigation. This illustrates that our module can dynamically bias towards more important modalities at the right time to learn a state representation (i.e., type of ore) useful for completing the task.
	
	In addition, we use the Perturbation-based saliency \cite{greydanus2018visualizing} for generating saliency maps, and use it to illustrate what features the agent pays attention to and whether agents are making decisions for the right reasons. As shown in Figure \ref{fig:analysis2}, we plot a sequence of frames with saliency maps in the Mining domain. We can see that the agent exhibits a significant change in their attention at step 2, as the agent is currently near the ore and know its type through the audio (i.e., the agent should pick the stove). 
	This shows that the agent can find relationships and correspondences between sub-components of instances from multiple heterogeneous modalities, i.e., it focuses its attention on the corresponding tool when it hears a sound at step 2.
	The experiment further analyzes that our module can dynamically bias towards more important modalities at the right time and make decision for the right reasons.
	
	\subsection{Case Study: Self-Driving Car Control}
	
\begin{figure}[t]
		\centering
		\includegraphics[width=0.98\columnwidth]{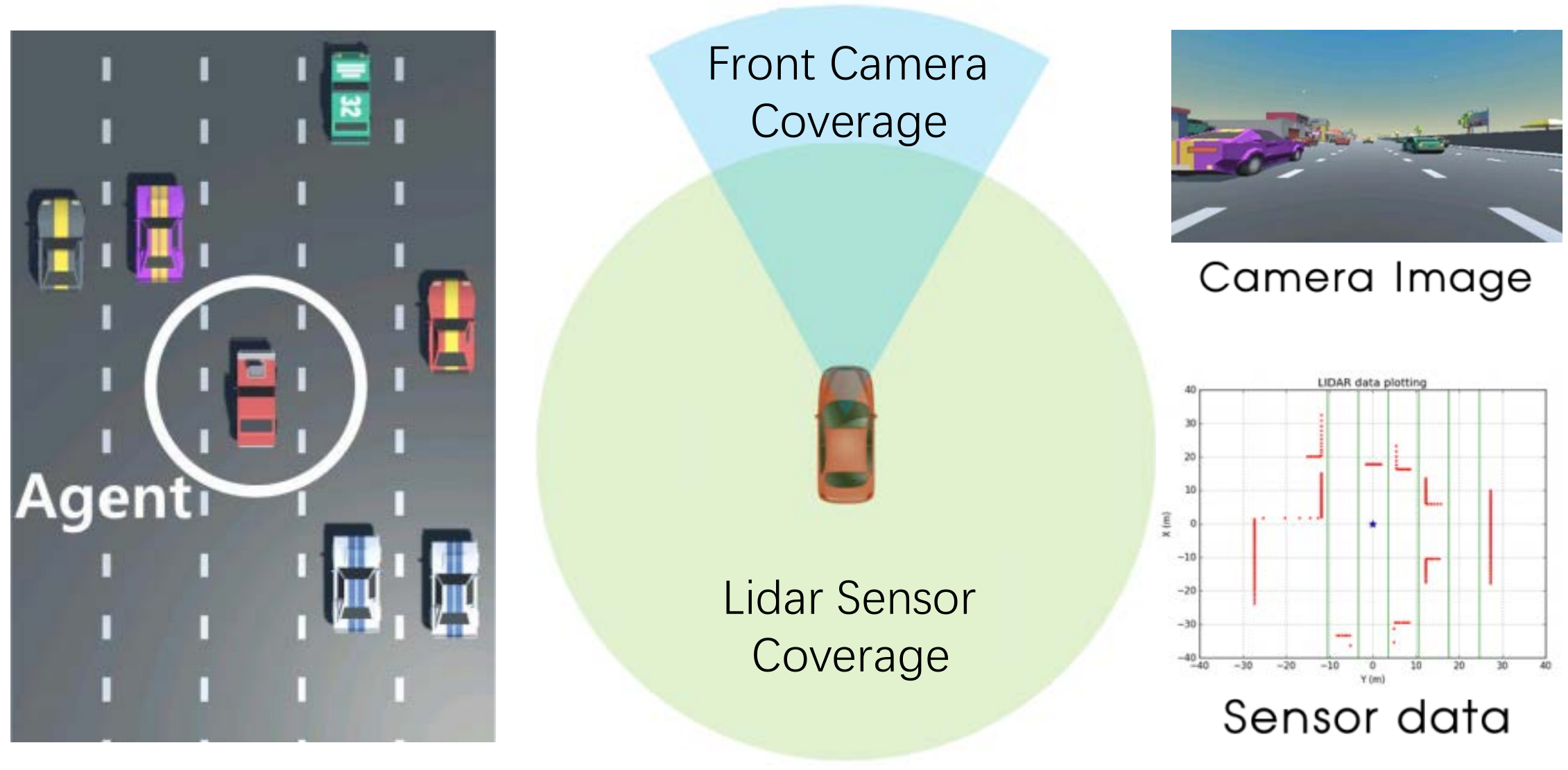}
		\caption{Highway and sensor configuration.}
		\label{fig:driving}
	\end{figure}

	\noindent To test our techniques in more realistic environments, we use a simulator for self-driving car control \cite{min2018deep}, which is developed for research on RL agent with multi-sensory inputs. As shown in Figure \ref{fig:driving}, the agent must drive safely on a highway composed of five lanes and avoid collisions with other vehicles. It has two inputs: camera image and lidar data. Specifically, there is a camera on the front to capture objects that are far ahead, and a raw pixel image is observed for every step. A lidar sensor detects the range of 360 degrees that has one ray per degree, so it can capture objects around the vehicle. As the camera image and lidar data come from different modalities, the agent needs to leverage their respective advantages to  maximizes the expectation value of the future reward. Here, the high reward is composed of high speed, low frequency of lane changes and collision-free. Notice that the simulator used in our experiments has already provided sufficient features for multi-modal RL and has been used by others in the literature \cite{min2018deep}. More sophisticated simulators may offer additional features such as user scenario definition and different car control models. However, they are not fundamental for evaluating the performance of different multi-modal RL algorithms.
	
	Our results are summarized in Table 1. As expected, the average reward of Image-Lidar (i.e., simply concating image and lidar data) has a higher value of 2457.21 than Image-Only (1997.12) and Lidar-Only (2153.20), which confirms the benefit of multimodal learning. Compared to lidar, camera is less reliable and gets the lowest reward in worst case and also higher deviation. For example, a sudden lane change may cause a collision if the agent cannot observe the approaching vehicles behind. However, directly combining image and lidar data (i.e., Image-Lidar (397.75)) makes such situation even worse than Lidar-Only (508.53) though better than Image-Only (202.89). As aforementioned, the reason is that learning state representation from multimodal inputs is challenging.
	
	Our method substantially outperformed all the compared methods, and achieved the best average reward (2611.25), the lowest deviation (556.99), and the best reward in worst case (920.18). Most importantly, our method is 80.9$\%$ higher than Lidar-Only in worst case. CASL did not achieve good results in the case study. As we mentioned before, the attention mechanism has more complex structure and requires a large number of samples to train. These results show the potential of our method to improve the safty issues of self-driving car when controlled by RL agent with multi-sensory inputs, thanks to our modality alignment and importance enhancement modules.
	
	\begin{table}[t]
		\centering
		\label{table1}
		\caption{Results of Self-Driving Car Control. The performance of all the methods is the average of three training runs with different random seeds after convergence.}
		\begin{tabular}{ccc}
			\toprule
			{\bf Method} & {\bf Average} & {\bf Worst-Case} \\
			\midrule
			MAIE  & \textbf{2611.25 $\pm$ 556.99} & \textbf{920.18}\\ 
			CASL  & 2144.52 $\pm$ 660.24 & 158.82 \\
			Image-Lidar  & 2457.21 $\pm$ 876.57 & 397.75\\ 
			Image-Only  & 1997.12 $\pm$ 749.67 & 202.89\\ 
			Lidar-Only  & 2153.20 $\pm$ 642.75 & 508.53\\
			\bottomrule
		\end{tabular}
	\end{table}

	\section{Conclusions}
	
	\noindent In this paper, we addressed the two major challenges of training RL with multimodal information, i.e., the heterogeneity and dynamic importance of different modalities. Specifically, we proposed a novel multimodal RL approach with the modality alignment and the importance enhancement modules, targeting the two challenges respectively. In the experiments, we compared with several state-of-the-art methods and show the advantage of our approach in terms of training speed and solution quality. Through module analysis, we demonstrate that: 1) our method can properly align features of different modalities along with the underlying state, and 2) dynamically adjust the weights of modalities based on their importance for representing the state. To put them together, we contribute to the community with a novel and effective approach for learning state representations in multimodal RL. Other than the challenges considered here, it is also interesting to explore multimodal RL with inconsistent or even conflict information in different modalities, which is left for future work.

	\bibliographystyle{IEEEtran} 
	\bibliography{main}

	\vfill
	
\end{document}